\documentclass{llncs}

\input{psfig.sty}
\usepackage{graphicx}
\usepackage{epstopdf}
\usepackage{lipsum}
\usepackage[table]{xcolor}
\usepackage{subfigure}
\usepackage{array}
\newcolumntype{P}[1]{>{\centering\arraybackslash}p{#1}}

\begin{document}



\title{A Markov Random Field  and Active Contour Image Segmentation Model for Animal Spots Patterns}


\author{Alexander G\'omez\inst{1} \and German D\'iez\inst{1} \and Jhony Giraldo\inst{1} \and Augusto Salazar\inst{1,}\inst{2} \and Juan M. Daza\inst{1}}

\institute{Universidad de Antioquia UdeA, Calle 70 No. 52 -- 21, Medell\'in, Colombia.\and  Instituto Tecnol\'ogico Metropolitano ITM, Carrera 21 No. 54-10, Medell\'in, Colombia.\\
\email{\{alexander.gomezv,german.diezv,
heriberto.giraldo\}@udea.edu.co }}


\maketitle

\begin{abstract}

Non-intrusive biometrics of animals using images allows to analyze phenotypic populations and individuals with patterns like stripes and spots without affecting the studied subjects. However, non-intrusive biometrics demand a well trained subject or the development of computer vision algorithms that ease the identification task. In this work, an analysis of classic segmentation approaches that require a supervised tuning of their parameters such as threshold, adaptive threshold, histogram equalization, and saturation correction is presented. In contrast, a general unsupervised algorithm using Markov Random Fields (MRF) for segmentation of spots patterns is proposed. Active contours are used to boost results using MRF output as seeds. As study subject the  \textit{Diploglossus millepunctatus} lizard is used. The proposed method achieved a maximum efficiency of $91.11\%$.

\end{abstract}


\section{Introduction}
Animal biometrics has increased in recent years, identifying individual animals and recognizing them at different places and time is an important requirement in many biological tasks like calculating animal population density, survival, emigration, examination of a particular behavior and planning conservation measures~\cite{kuhl2013animal}. Commonly applied strategies can be categorized in two classes: intrusive and non-intrusive. Intrusive approaches include marking animals, which involves capture and risk the animal to injury, modify its behavior and even changes survival possibilities~ \cite{kelly2001computer}; also marking strategies are not suitable in large populations or for long time. Non-intrusive approaches include the identification of genetic markers in excrement~\cite{lahiri2011biometric} and photographic mark recapture (PMR)~\cite{bolger2012computer}. 

The PMR method is based on visual identification using phenotypic features like spots, stripes or morphology. Those features must be stable over time, unique, and photographed under different conditions. This method is a two photo comparison of one target and hundreds of possible subjects to test similarity between patterns. For this reason large animal populations impede the identification by a human observer, because subjectivity, skill or experience of the expert would affect the objectivity of the study~\cite{kuhl2013animal}.

Automatic biometric identification (possible in PMR) is a time-saving alternative that provides clearness to the identification process. Previous semi-automatic approximations include shapes for marine mammals~\cite{gope2005affine}, elephants, and some lizard species; stripes for zebras ~\cite{lahiri2011biometric} and tigers; also spots for cheetahs, giraffes~\cite{bolger2012computer}, marine turtles and polar bears. There are two possible scenarios for a computer vision perspective. First, photos taken in the wild as photo trap framework; this media is commonly cluttered, with low contrast, containing trees, shrubs, other subjects, and the target in multiple poses~\cite{ardovini2008identifying}. Second, the subject is photographed under controlled conditions and position. Additionally to the scenario, both cases present problems in natural appearance of skin, brightness, 3D shape, contamination produced by sand or environmental components, and scars. Due to the spot concept that is related to a contrast change between two or more regions, the purpose of this paper is to find a general algorithm for segmentation, whatever kind of spots set on animals that deal with the previous declared problems. A general algorithm for spots patterns give the opportunity to identify a great variety of animals, e.g., the above mentioned salamanders and whale sharks.

Our study subjects are the endangered lizards \textit{Diploglossus millepunctatus} from Malpelo Island (Colombia)~\cite{lopez2006lizards}. These reptiles present an unique spot pattern per subject and currently are studied using mark based methods. Due to the structured scales comprising the lizard's skin it has 3D variations influencing the illumination. The spots have an non-uniform color distribution and can be blurred or highly defined, or occluded by residuals from food, garbage or excrement. All possible variations in the spot segmentation problem for animal biometrics are present in this scenario. 

In general, biometrics approaches analyze a region of interest (ROI), which is manually selected and then a segmentation is done giving seeds (also manually selected) to an adaptive shape algorithm as deformable shapes or active contours. We propose an automatic method for spot segmentation that avoids user initialization or seeds. Our results show that simple cost functions with MRF framework can perform powerful and effective segmentation of these patterns in multiple illumination variations and under noisy conditions. 

Related work is mentioned in Section~\ref{sec:related}. In Section~\ref{sec:methods} the methods used in the model are explained. Section~\ref{sec:experiments} describes the experiments used to test the model, and Section~\ref{sec:results} shows and discusses the results. Finally, in Section~\ref{sec:conclusions} conclusions and future work are presented.

\section{Related Work}\label{sec:related}

The \textit{Diploglossus millepunctatus} spots have no the same intensity values throughout the whole subject. This issue is most critical when high amounts of light irradiate the lizard and mask the spots in the illuminated regions. This issue can be modeled with a Markov Random Field (MRF) that can deal with uncertainty of pixel intensities that belong to a spot in a determinate region based on multiple soft criteria like local intensity, neighborhood relations and a broad number of patterns.

MRFs have been proven to be a suitable image model to resolve computer vision tasks like image segmentation. Boikov and Jolly~\cite{boykov2001interactive} showed that with some seeds set by the user any object can be segmented using hard constraints and histograms for object and background. In~\cite{rother2004grabcut} the histograms of user seeds were replaced by Gaussian Mixture Models (GMM), one for background and one for foreground, and also a border matting algorithm was developed to fix transparency on segmented object edges. Another approach is~\cite{kumar2005obj} where a shape model was imposed through Layered Pictorial Structures to MRF, which favored specific trained shapes (cows) but need user initialization and a training stage. The method in~\cite{kato2006markov} did not need user interaction or training, it is based on color values from CIE-L*u*v* color space and texture features from Gabor filtered images as data term, with a GMM parameterized automatically with EM algorithm. However, estimating the number of classes highly depends on the image appareance. In~\cite{delong2009globally} the authors propose a multi-region segmentation method based on geometric interactions between objects that were previously segmented with user interaction or automatic framework. Previous segmentation algorithms showed excellent performance but all of them need seeds or depend on image conditions.

Our approach uses monogrid model-based segmentation, does not need user interaction and targets a specific object (spots) in challenging scenarios, without previous training, using an appearance model based on RGB color space, gray-level image and smoothness constraints. Segmentation has been proven in hard light contamination conditions, noisy and blurry images with $3$ types of models and $2$ inference algorithms.

\section{Methods}\label{sec:methods}

\begin{figure}[hbtp]
\centering
\includegraphics[width=\textwidth]{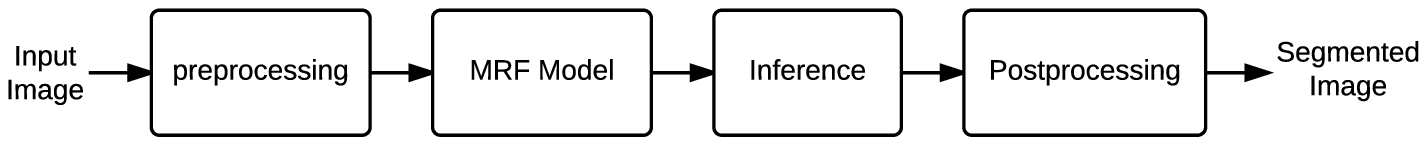} 
\caption{Scheme of the proposed method.}\label{fig:methods}
\end{figure}

Here the proposed method is described, as shown in Figure.~\ref{fig:methods}, where the method is subdivided in its main processes. The preprocessing step highlights characteristics and helps to enhance the model’s score. The MRF model block extracts parameters from the input image to feed the mathematical model, and the inference solves the maximum a posteriori probability problem of the MRF model and gives a mask with spots.

\subsection{Preprocessing}

\begin{figure}[hbtp]
\centering			
\subfigure[]{\includegraphics[width=0.2\textwidth]{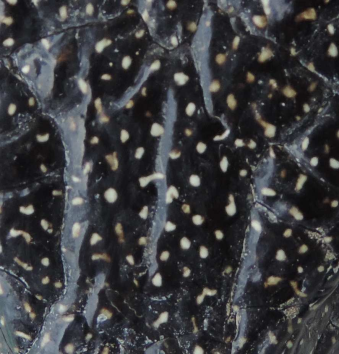}{\label{fig:preprocessing1}}}
\subfigure[]{\includegraphics[width=0.2\textwidth]{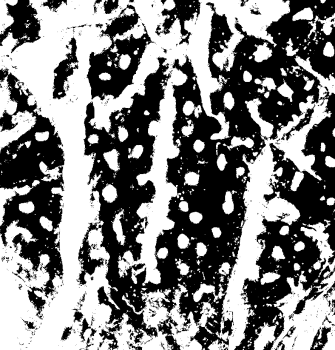}{\label{fig:preprocessing2}}}
\subfigure[]{\includegraphics[width=0.2\textwidth]{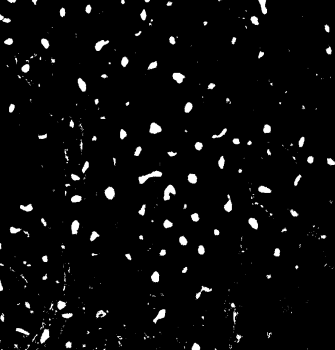}{\label{fig:preprocessing3}}}
\subfigure[]{\includegraphics[width=0.2\textwidth]{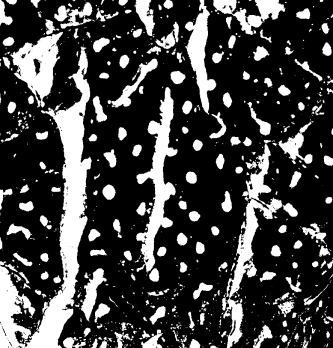}{\label{fig:preprocessing4}}}
\subfigure[]{\includegraphics[width=0.2\textwidth]{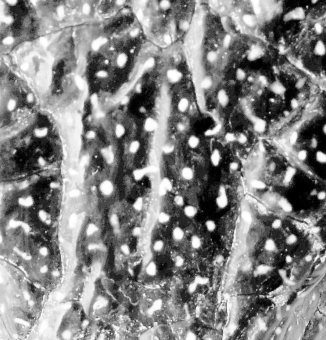}{\label{fig:preprocessing5}}}
\subfigure[]{\includegraphics[width=0.2\textwidth]{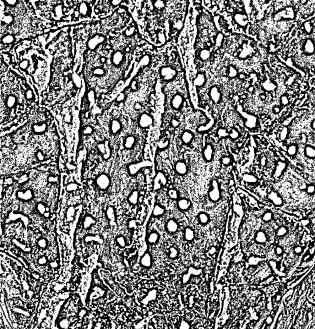}{\label{fig:preprocessing6}}}
\subfigure[]{\includegraphics[width=0.2\textwidth]{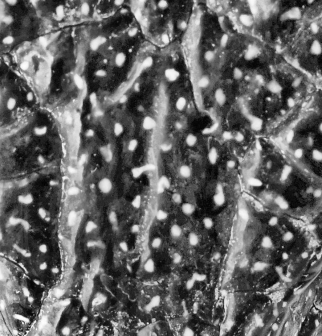}{\label{fig:preprocessing7}}}
\subfigure[]{\includegraphics[width=0.2\textwidth]{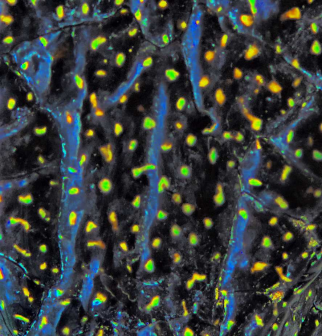}{\label{fig:preprocessing8}}}
\caption{(a) Raw image. (b) Low threshold value. (c) High threshold value. (d) Otsu’s threshold method. (e) histogram equalization. (f)adaptive threshold. (g) CLAHE (h)Preprocessed image.
A single method only cannot isolate the spots without a significant loss of spots or a noise introduction.}
\label{fig:preprocessing}
\end{figure}

Non-uniform illumination and non-constant color of \textit{Diploglossus millepunctatus} spots are essential objectives for preprocessing steps, since there is no threshold that can separate spots from foreground, a low value in binarization lets pass all the spots, but also large amounts of light (Figure~\ref{fig:preprocessing2}). Moreover a high threshold (Figure~\ref{fig:preprocessing3}) lets only pass the desired pattern, but misses low-intensity spots. There is no prior knowledge about the optimal threshold value on every image. A common solution is Otsu's method (Figure~\ref{fig:preprocessing4}), which assumes binarization like bi-class clustering problem and selects a threshold value that minimizes intra-class variation. 

Histogram preprocessing techniques to enhance contrast include histogram equalization and contrast correction. Histogram equalization is a global method that sparse the histogram of an image; however this approximation does not produce good results (Figure~\ref{fig:preprocessing5}), because it makes the spots closest to brightness regions intensities. Contrast correction is a point operation that enhances contrast multiplying intensities of a pixel by a fixed value between $1$ and $3$ and casting it to a value between $0$ and $255$, this causes a significant contrast enhancement in dark regions, but gaps among spots and higher intensity regions remain unchanged. Global techniques as histogram equalization or point operations like contrast correction are strategies that use global statistics of an image or just modified pixel values with a constant; they do not observe local variation on contrast and assume equal distribution of intensities in an image. 

Local operations like adaptive thresholding (AT) and contrast adaptive histogram equalization (CLAHE) observe a local window in each pixel and calculate the optimum threshold value or intensity to a split histogram. Local algorithms depend on window selections and since dark and bright regions, size and distributions, are aleatory, windowing size has to vary throughout the image. Results using AT and CLAHE are exhibited in Figure~\ref{fig:preprocessing6} and Figure~\ref{fig:preprocessing7}, both reflect bad choices of correction values, caused by the fixed size of the observed window.

The proposed method  equalizes light and keeps the color values constant to exploit spot color information, this reasoning is done using color spaces that convert RGB color space to representations independent of brightness, HSV, L*a*b* and HSI color spaces are representations that deal with this problem. Due to the equalization of aleatory light distribution a CLAHE was applied in the brightness channel on 3 spaces, thus the L*a*b* space shows a more uniform distribution. In order to separate spots from light regions, a saturation correction was implemented. This process highlights Red and Green channels; hence spots were turned brighter than the light regions (seeFigure~\ref{fig:preprocessing8}).

\begin{figure}[hbtp]
\centering
\includegraphics[width=\textwidth]{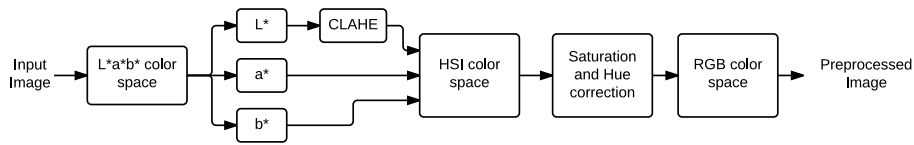}
\caption{Schema of the proposed preprocessing method}\label{fig:preprocessingE}
\end{figure}

Figure~\ref{fig:preprocessingE} shows the proposed preprocessing method applied to the input images. First, local correction (CLAHE) in the Luminance channel of L*a*b* space is applied, followed by a point operation (saturation correction) in HSI color space, and finally the image is transformed to RGB space.

\subsection{MRF Model for segmentation}

Image segmentation ideology assumes that a scene consists of a finite number of regions with characteristics, which change slowly and could be identified with the image constitutive elements. The segmented image is a simplification of the source image where every identified region has a label that classifies them into an image feasible class.

Several prominent traditional segmentation algorithms have been based on probabilistic graphical methods, in which there are sets of observed random variables, hidden random variables and observations over some random variables. Probabilistic approaches try to calculate the probability of a pixel or number of pixels belonging to a certain feasible image class. These classes are discrete random variables, taking values in $L = \{1, 2, .., L\}$, with L as the maximum number of feasible classes in the image. The set of these labels is a random field, called the label process~\cite{kato2006markov}. In this approach, the random variables are related through energy functions that determine whether the pixel belongs to a determined class. The inference process is based on both, the individual values of the pixel or group of pixels and the neighborhood relations. These relations are computed using cliques~\cite{koller2009probabilistic}. This graph topology allows the interaction of each pixel or group of pixels only with their closer neighborhood, which is called a first order Markov blanket.

One form to define an energy function $E$ is to define it in terms of the disagreement between the observed data or $E_{data}$ and the measurement of the extent to which $E$ is not piecewise smooth or $E_{smooth}$, such as $E = E_{data} + E_{smooth}$. The selection of the energy functions is a difficult task because different elections in the $E_{smooth}$ and $E_{data }$ produce different results in the final segmented image. $E_{data}$ term could consider different factors as the interaction with the user, the shape or the different characteristics of the target object~\cite{lezoray2012image}. In non-supervised object segmentation the introduction of previously known information about the target object to the energy function as foreground specific intensity range of values or background-foreground contrast information could even improve the inference process.


In this work, three different energy functions were tested in order to properly represent the task to solve in the segmentation process. In those energy functions $I_{Gp}$ represents the gray-scale intensity value of the pixel $p$. Table~\ref{tab:energies} shows selected energy functions.

\begin{table}[h]
\centering
\setlength{\tabcolsep}{6pt}

\caption{Energy functions}
\begin{tabular}{clll}
\hline
\textit{Function} & $E_{data}^1$  & $E_{data}^0$ & $E_{smooth}$ \\
\hline
1          &    $\sum_{p\epsilon P}250-I_{Gp}$    & $\sum_{p\epsilon P}I_{Gp}$      &  Potts\\
2          &      $\sum_{p\epsilon P}250-I_{Gp}$   & $\sum_{p\epsilon P}I_{Gp}*I_{Lp}$      &  Potts\\
3          &        $\sum_{p\epsilon P}250-I_{Gp}$  & $\sum_{p\epsilon P}I_{Gp}*I_{c}$      &  Potts\\
\hline
\end{tabular} \label{tab:energies}
\end{table}

\textit{Function 1} approaches the problem based on a priori knowledge of the spot structure, i.e, high values in grayscale values and lower values of background. Since spots and the background vary in the image, some possible spots regions will have more probability in light regions than dark ones. Also, the gap energy between spots and background regions is higher in dark regions. \textit{Function 2} uses a grayscale image discretized to $2$ levels. Assuming that spots must become to higher values and background to lower ones, this knowledge is incorporated through the binary variable $I_{Lp}$, which takes value $1$ when $I_{Gp}$ is $255$ or $0$ otherwise. Finally, \textit{Function 3} uses previous information about the spot color histogram, using a binary weight $I_{c}$ with value $1$ when red and green channels in the input image have an intensity lower than the blue one, and $0$ otherwise.

A common smooth term energy function is the Potts model, which is the simplest discontinuity preserving model, where discontinuities between any pair of labels are penalized equally and can be reduced to the multi-way minimization problem~\cite{boykov2001interactive}, which is known to be a NP-complete problem where NP means non-deterministic polynomial time.

In order to solve the inference task, this work uses two algorithms: Graph Cuts (GC) using a push relabel approach and Loopy Belief Propagation (LBP); each algorithm has a different approximation to the solution, and thus, the final results differ, too.

\subsection{Postprocessing}

Active Contours are energy-minimizing spline curves, guided by external constraint forces and influenced by image forces that pull it towards features such as lines and edges~\cite{kass1988snakes}. The internal forces in a spline curve impose smoothness constraints. The image forces moves the \textit{snake} toward salient image features like edges and lines. Finally, external constraints put the \textit{snake} in a local minimum. The energy of the \textit{snake} can be written as a sum of $E_{int}$ and $E_{image}$. $E_{int}$ term controls elasticity and stiffness, and $E_{image}$ uses image features to reduce the energy of the expression, e.g. edges and gradients. This energy is minimized by iterative algorithms. Because Active Contours require a seed to begin MRF segmentation, results are used to initialize the algorithm and to enhance the performance. Active contours have a parameter called contraction bias that is part of $E_{int}$ and the best value was found through experimentation and set to $-0.4$.


\section{Experimental framework}\label{sec:experiments}

\subsection{Dataset}

\begin{figure}[hbtp]
\centering
\includegraphics[scale=0.3]{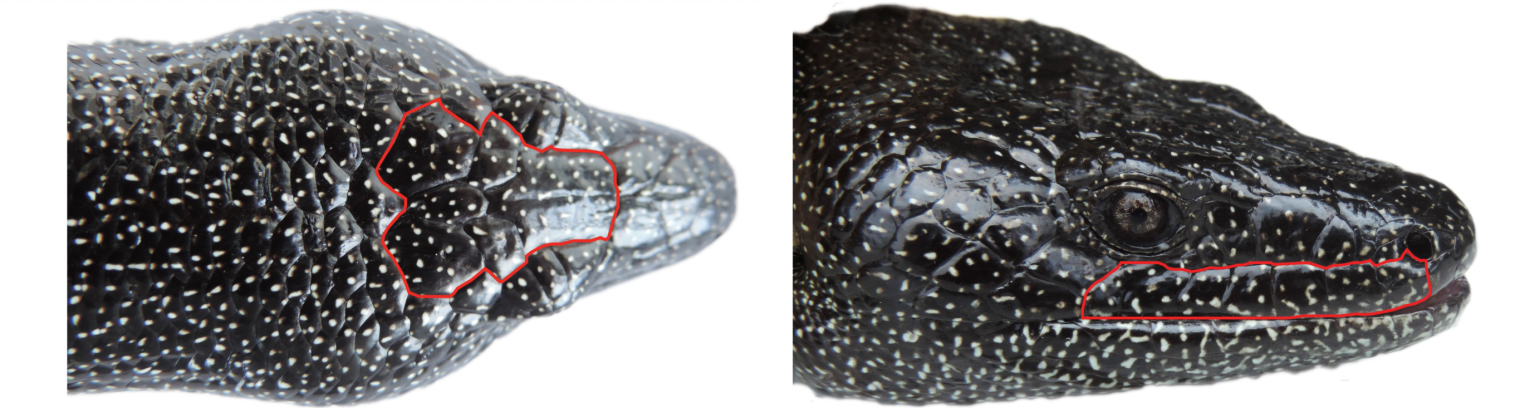}
\caption{ROIs from the \textit{Diploglossus millepunctatus} lizard.}
\label{fig:flakes}
\end{figure}

The database used in this work was provided by the Exact and Natural Science Department of the University of Antioquia and consists of images from 19 individuals taken under controlled conditions. The database is limited, because the ground truth must be obtained by an expert segmenting manually each image, which is quite time-consuming. As is usual in visual expert identification, two homologous regions are selected in order to perform an identification between individuals as shown in Figure~\ref{fig:flakes}. Five images from each individual were taken, 3 frontal images and 2 lateral images, and with this images the expert obtains the ground truth.

\subsection{Experiments}

Based on the methods previously explained, a set of experiments was planned, in order to find the combination that provides the best segmentation. Table~\ref{tab:experiments} lists the experiments performed where the energy functions described in Table~\ref{tab:energies} are proved to find which one fits more to the problem. Preprocessing and postprocessing stages were used to enhance  thhe functions performance and reduce the final error, respectively. 

\begin{table}[hbtp]
\centering
\setlength{\tabcolsep}{6pt}
\caption{Description of the experiments}
\begin{tabular}{lllll}
\hline
Test & Preprocessing  & Energy & Inference & Postprocessing \\
\hline
Exp1          &      None             & \textit{Function 1}      & LBP/GC &  None\\
Exp2          &      None             & \textit{Function 2}      & LBP/GC & None\\
Exp3          &      None             & \textit{Function 3}      & LBP/GC &  None\\
Exp4          &      None             & \textit{Function 1}      & LBP/GC &  Active Contours\\
Exp5          &      None             & \textit{Function 2}      & LBP/GC &  Active Contours\\
Exp6          &      None             & \textit{Function 3}      & LBP/GC &  Active Contours\\
Exp7          & Proposed 	  	 & \textit{Function 1}      & LBP/GC &  None\\
Exp8          & Proposed       & \textit{Function 2}      & LBP/GC &  None\\
Exp9          & Proposed 	  	 & \textit{Function 3}      & LBP/GC &  None\\
Exp10         & Proposed 	 	 & \textit{Function 1}      & LBP/GC &  Active Contours\\
Exp11         & Proposed 	 	 & \textit{Function 2}      & LBP/GC &  Active Contours\\
Exp12         & Proposed 	 	 & \textit{Function 3}      & LBP/GC &  Active Contours\\
\hline
\end{tabular}\label{tab:experiments}
\end{table}

Based on the ground truth, two different formal metrics, confusion matrix and Hoover metrics, were implemented in order to measure the segmentation performance. The confusion matrix compares the ground truth with the machine segmented image and weighs the percentage of pixels matched and mismatched based on the total number of pixels. Hoover metrics~\cite{hoover1996experimental} consider five types of regions in the ground truth and machine segmented image comparison, either classified as  correctly detected, over-segmented, under-segmented, missed and noise, and then plots the number of areas in each class weighted by total amount of areas based on a threshold (tolerance \%) term that is the free term in which the graphics are based.

\section{Results}\label{sec:results}

Table~\ref{tab:resultmat} shows the performance of the model with each cost function and inference algorithm, and the performance of each model after Active Contours were applied. The values correspond to the mean of efficiency in each condition, where efficiency is calculated as sum of true positive and true negative terms from the confusion matrix.

\begin{table}[hbtp]
\centering
\setlength{\tabcolsep}{6pt}
\caption{Efficiency of the segmentation for each of the experiments performed}
\begin{tabular}{P{1cm}P{1cm}P{1.3cm}P{0.5cm}P{1cm}P{1cm}P{1.3cm}P{0.1cm}}
\hline
Test & LBP  & GC&  &Test & LBP  & GC& \\
\hline
\centering
Exp1  &  71.86 & 71.92 & &Exp7     & 84.52 & 84.64  &\\
Exp2  &  66.75 & 66.56 & &Exp8     & 66.75 & 66.75 \\
Exp3  &  78.57 & 78.62 & &Exp9     & 82.48 & 82.54   \\
Exp4  &  78.52 & 84.91 & &Exp10    & 88.97 & 88.38    \\
Exp5  &  54.58 & 54.58 & &Exp11    & 64.64 & 64.64  \\
Exp6  &  91.11 & 90.23 && Exp12 & 90.21 & 90.2  \\
\hline
\end{tabular}\label{tab:resultmat}
\end{table}



The results show that a cost function built with intensity differences \textit{Function 1} performs bad per-pixel segmentation when the image has low contrast between foreground and background. However preprocessing enhances this performance significantly pushing the efficiency from 71.86\% to 84.52\%  raising the contrast gap and reducing false positives. \textit{Function 2} showed the worst results due to insufficient seed provision. The color-based cost function \textit{Function 3} shows the best results in raw images, owing to the color nature of lizard spots; in preprocessed images it reaches similar results to \textit{Function 1}. Using active contours, with MRF as seeds, enhanced the results up to $91.11\%$ in raw images and $90.21\%$ in preprocessed images, because active contours correct under-segmented instances augmenting regions from the seed.

\begin{figure}[hbtp]
\centering			
\subfigure[]{\includegraphics[width=0.48\textwidth]{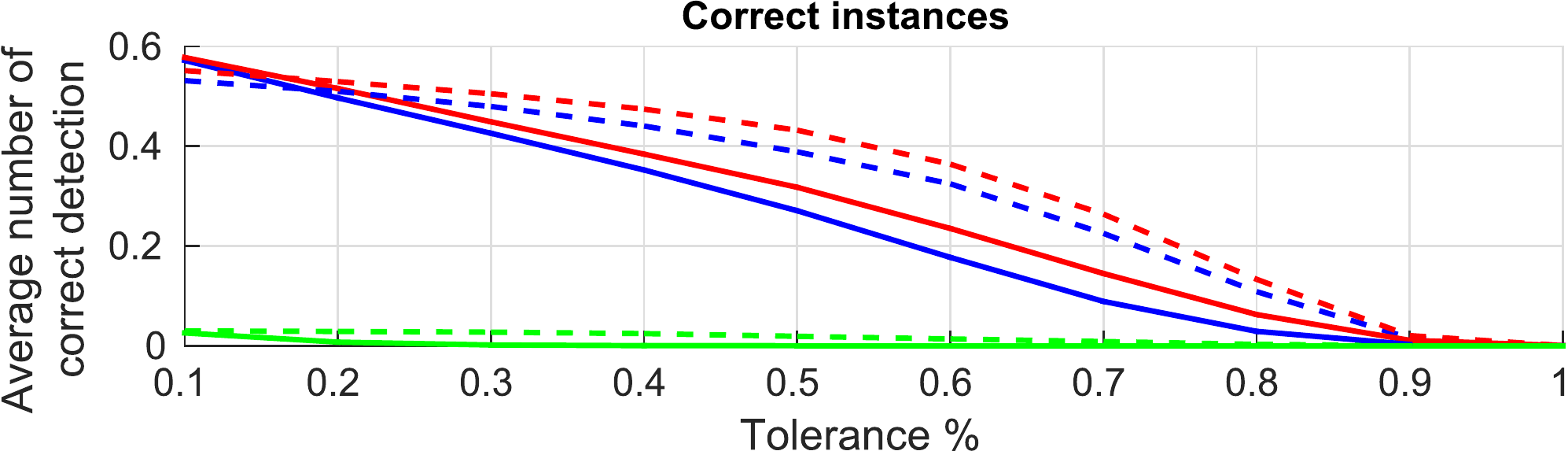}{\label{fig:hoover1}}}
\subfigure[]{\includegraphics[width=0.48\textwidth]{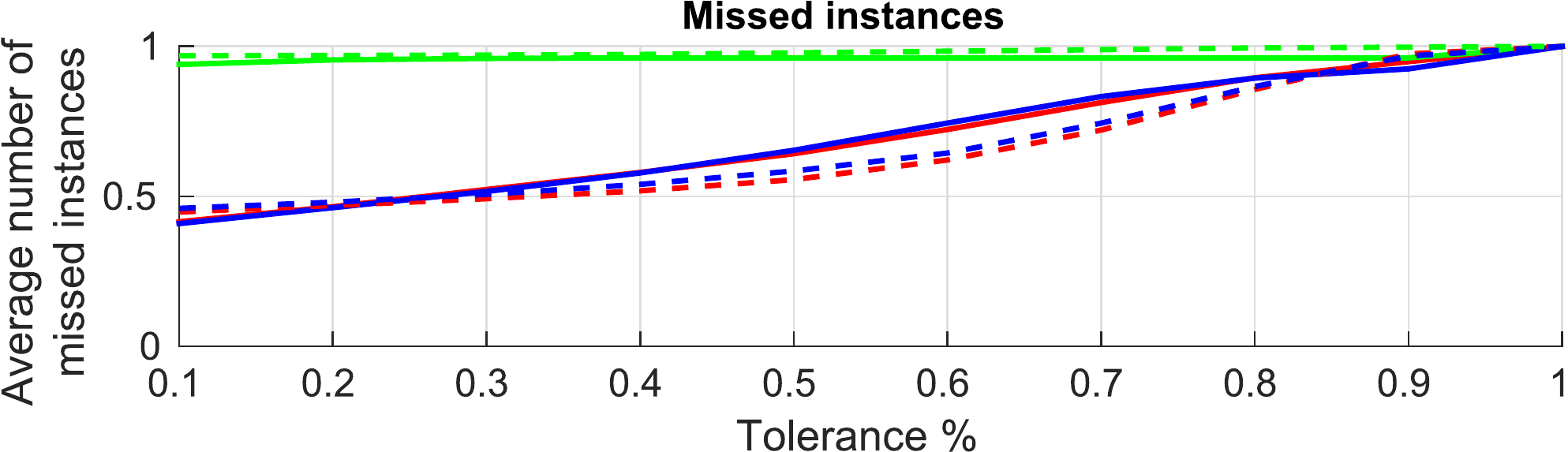}{\label{fig:hoover2}}}
\subfigure[]{\includegraphics[width=0.48\textwidth]{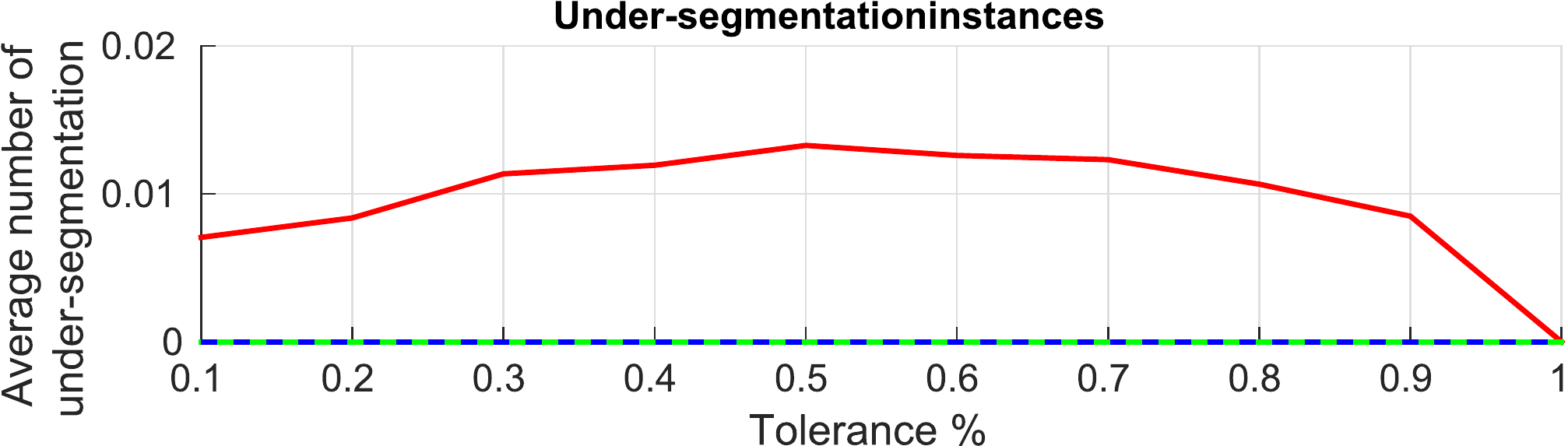}{\label{fig:hoover3}}}
\subfigure[]{\includegraphics[width=0.48\textwidth]{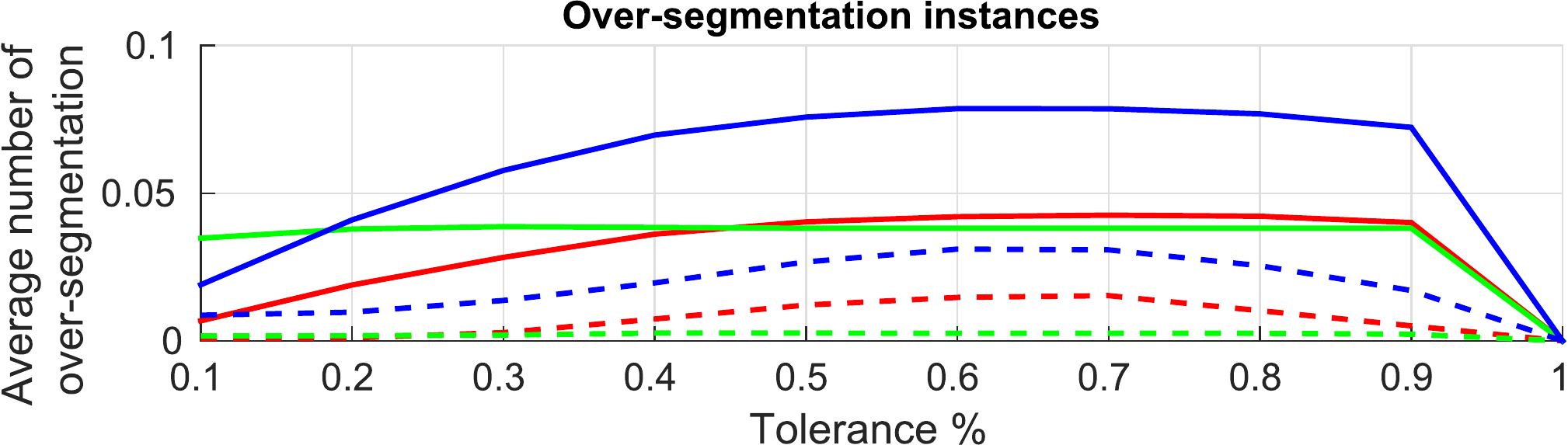}{\label{fig:hoover4}}}\\
\subfigure[]{\includegraphics[width=0.61\textwidth]{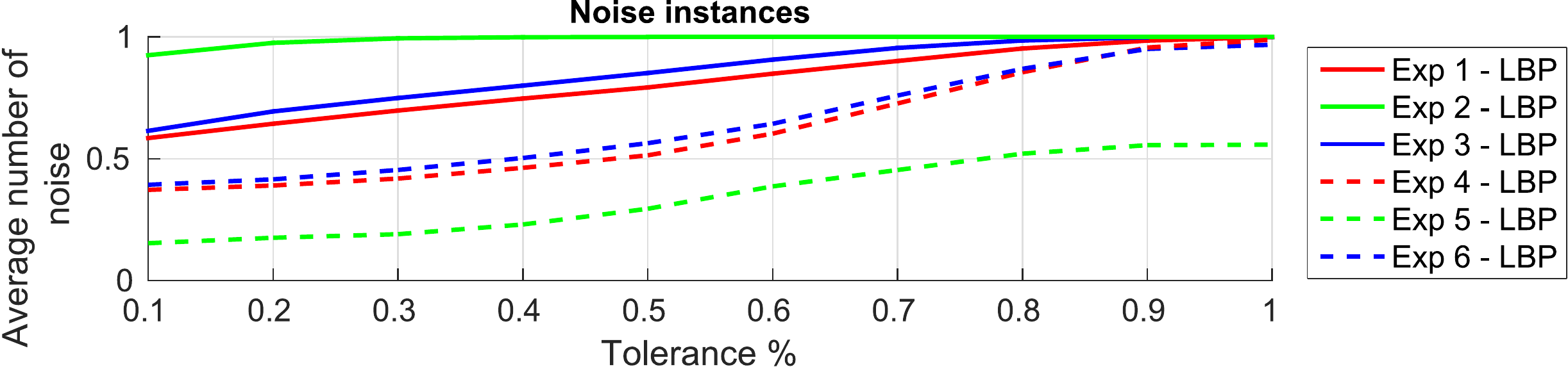}{\label{fig:hoover5}}}
\caption{Average of the Hoover metrics. (a) Correct instances. (b) Missed instances. (c) Under-segmented instances. (d) Over-segmented instances. (e) Noise instances}
\label{fig:hoover}
\end{figure}

Since confusion matrices do not expose segmentation quality and just give an idea of correct identified pixels, Hoover metrics~\cite{hoover1996experimental} gives a wide intuition of the method performance and are presented in Figure~\ref{fig:hoover} just LPB inference is exposed due to inference algorithms produce slightly differences in the graphics.

Hoover metrics shows that \textit{Function 1} produces the better region performance in all metrics proving that color information is not determinant for good segmentation. The Figures~\ref{fig:hoover1} and \ref{fig:hoover2}  exhibit that the intensity based function has problems delimiting spots. This probably is caused by the short gap between spots and background in light regions. Figure~\ref{fig:hoover4} shows how non-uniform color distribution inside spots causes over-segmentation in color based on the energy \textit{Function 3}.

Figure~\ref{fig:hoover3} and Figure~\ref{fig:hoover4} demonstrate that the models does not suffer meaningful under-over segmentation problems, giving less than $2\%$ and $10\%$, respectively. The Active Contours enhance over-segmented images merging regions that are inside a spot in the input image and adjust to the original spot of the input image in under-segmented images. Noise regions (see Figure~\ref{fig:hoover5}) are reduced with Active Contours, because the input image does not have a valid region's contour to adapt. In contrast, missed regions increment when a spot in the input image does not have a visual region to adapt.

\begin{figure}[hbtp]
\centering		
\subfigure[]{\includegraphics[width=0.2\textwidth]{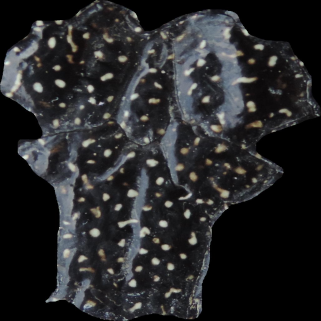}{\label{fig:resul_gt1}}}
\subfigure[]{\includegraphics[width=0.2\textwidth]{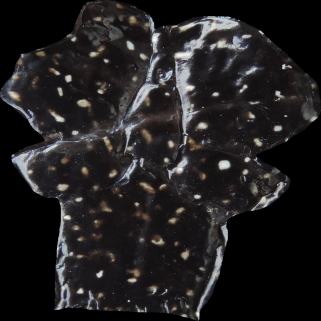}{\label{fig:resul_gt2}}}
\subfigure[]{\includegraphics[width=0.2\textwidth]{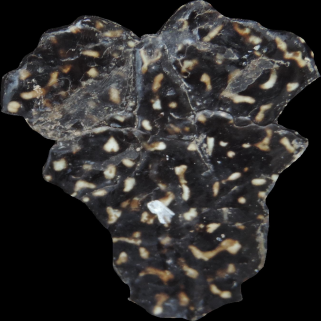}{\label{fig:resul_gt3}}}
\subfigure[]{\includegraphics[width=0.2\textwidth]{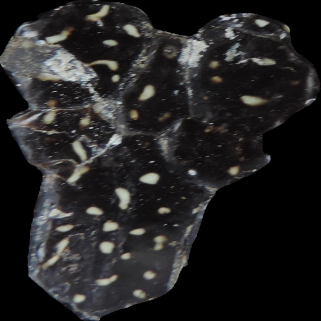}{\label{fig:resul_gt4}}}
\subfigure[]{\includegraphics[width=0.2\textwidth]{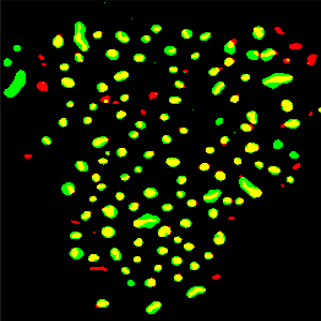}{\label{fig:resul_gt5}}}
\subfigure[]{\includegraphics[width=0.2\textwidth]{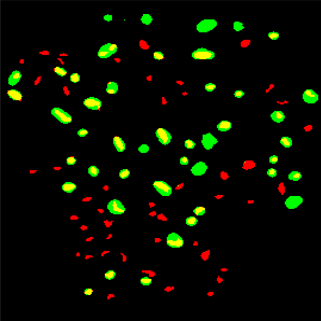}{\label{fig:resul_gt6}}}
\subfigure[]{\includegraphics[width=0.2\textwidth]{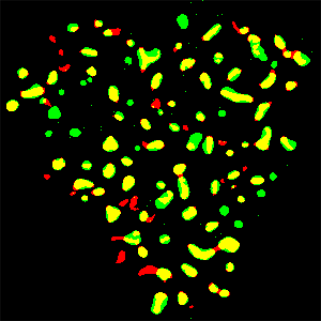}{\label{fig:resul_gt7}}}
\subfigure[]{\includegraphics[width=0.2\textwidth]{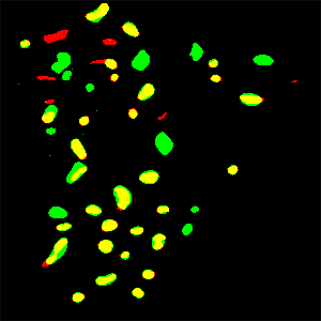}{\label{fig:resul_gt8}}}
\caption{Qualitative segmentation results. Raw images (upper row). Output images (lower row)}
\label{fig:resul_gt}
\end{figure}

To give a wide insight into the algorithm performance Figure~\ref{fig:resul_gt} compares the ground truth and the machine segmented image; Figures~\ref{fig:resul_gt1} to \ref{fig:resul_gt4} are input images. The Figures~\ref{fig:resul_gt5} to \ref{fig:resul_gt8} are the output of the MRF with Active Contours postprocessing. Pixels the model identifies as spots are green, but do not appear as a spot in the ground truth. Red pixels are ground truth spot pixels the model did not catch and yellow regions mean zones where the model accords with the ground truth.  The proposed model can solve the spot segmentation task under inner image variant illumination conditions, nevertheless, it is common that the model ignores spots that only span few pixels and also dark spots with intensity similar to the background. Large spots with narrow parts are usually divided into two parts with the narrow part as the break point, as common in energy segmentation approaches.

\section{Conclusions}\label{sec:conclusions}

In this paper a segmentation model for spots on animals based on Markov Random Fields and Active Contours is proposed and tested on \textit{Diploglossus millepunctatus} lizard images. Extensive experiments using energy functions ba\-sed on pixel intensities, quantization, and color information as cost functions were carried out. Also two inference methods, loopy belief propagation and Graph cuts were tested. A preprocessing approximation dealing with color spa\-ces, global and local enhancing, and segmentation methods was performed. The best performance was achieved with an intensity build data term function that reached $84.52\%$ using proposed preprocessing stage. Using Active Contours as postprocessing boosts the results up to $91.11\%$. The model shows promising performance to automatize segmentation processes in photographic mark recapture and to reduce processing time and subjectivity.

\begin{figure}[-h]
\centering			
\subfigure[]{\includegraphics[width=0.22\textwidth]{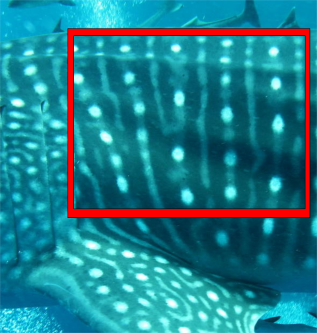}{\label{fig:resultshark1}}}
\subfigure[]{\includegraphics[width=0.23\textwidth]{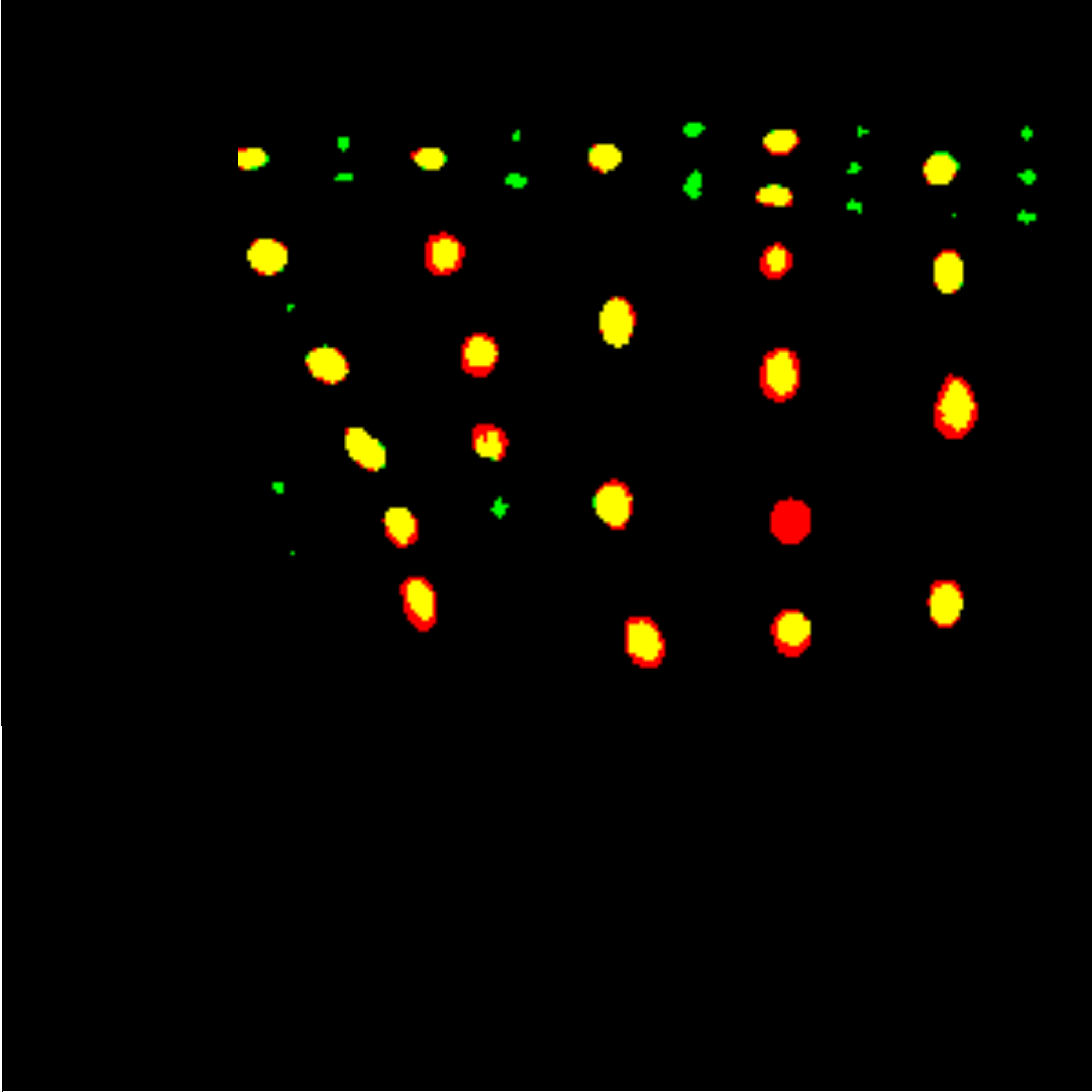}{\label{fig:resultshark2}}}
\subfigure[]{\includegraphics[width=0.2\textwidth]{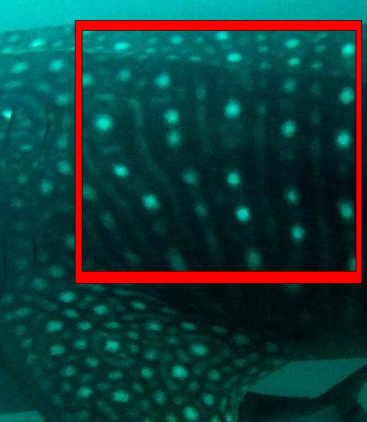}{\label{fig:resultshark3}}}
\subfigure[]{\includegraphics[width=0.23\textwidth]{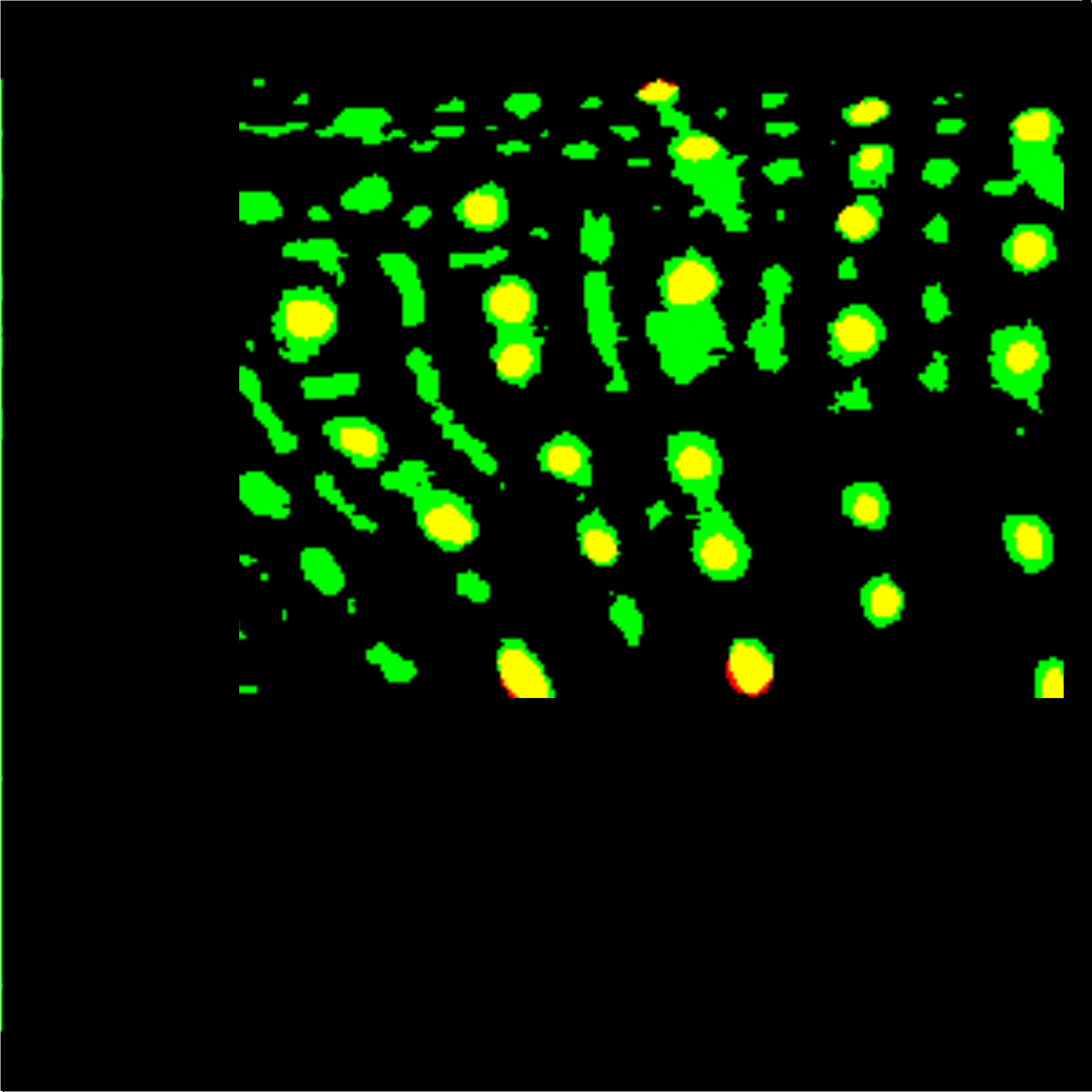}{\label{fig:resultshark4}}}
\caption{Result on whale shark database. (a) Input image 1. (b) Segmented image 1. (c) Input image 2. (d) Segmented image 2}
\label{fig:resultshark}
\end{figure}

In future work, the cost functions will have extra terms that include considerations of shape through a pictorial structures concept~\cite{kumar2005obj}. Color constrains will be modeled through GMM framework training and specifically modeled to \textit{Diploglossus millepunctatus} spots. The work will be extended to other animals and species. For instance, Figure~\ref{fig:resultshark} shows the cost function based on intensity applied to some samples from a whale shark dataset~\cite{holmberg2009estimating} to extract spot patterns. In this dataset, just one region of the whale shark is needed for identification (marked inside a red rectangle as in Figures~\ref{fig:resultshark1} and ~\ref{fig:resultshark3}). Figures~\ref{fig:resultshark2} and ~\ref{fig:resultshark4} show qualitative segmentation that follows the same notation from Figure~\ref{fig:resul_gt}. These results show promising performance assuming that the model segments all spots in the image without any additional tuning procedure. 

\bibliographystyle{unsrt}
\bibliography{isvc_submission}

\end{document}